\documentclass[letterpaper]{article} 
\usepackage{aaai24}  
\usepackage{times}  
\usepackage{helvet}  
\usepackage{courier}  
\usepackage[hyphens]{url}  
\usepackage{graphicx} 
\usepackage{natbib}  
\usepackage{caption} 
\usepackage{blindtext}
\usepackage{tabularx}
\usepackage{algorithm}
\usepackage{algorithmic}

\usepackage{newfloat}
\usepackage{listings}
\DeclareCaptionStyle{ruled}{labelfont=normalfont,labelsep=colon,strut=off} 
\floatstyle{ruled}
\newfloat{listing}{tb}{lst}{}
\floatname{listing}{Listing}
\title{Should agentic conversational AI change how we think about ethics? Characterising an interactional ethics centred on respect}
\author{
    Lize Alberts\textsuperscript{\rm 1}\textsuperscript{\rm 2}\textsuperscript{\rm 3},
    Geoff Keeling\textsuperscript{\rm 1},
    Amanda McCroskery\textsuperscript{\rm 1}
}
\affiliations{
    \textsuperscript{\rm 1}Google Research\\
    \textsuperscript{\rm 2} University of Oxford\\
    \textsuperscript{\rm 3} Stellenbosch University\\

    lize.alberts@cs.ox.ac.uk, gkeeling@google.com, amccroskery@google.com
}

\begin{document}

\maketitle

\begin{abstract}
With the growing popularity of conversational agents based on large language models (LLMs), we need to ensure their behaviour is ethical and appropriate. Work in this area largely centres around the `HHH' criteria: making outputs more helpful and honest, and avoiding harmful (biased, toxic, or inaccurate) statements. Whilst this semantic focus is useful when viewing LLM agents as mere mediums or output-generating systems, it fails to account for pragmatic factors that can make the same speech act seem more or less tactless or inconsiderate in different social situations. With the push towards agentic AI, wherein systems become increasingly proactive in chasing goals and performing actions in the world, considering the pragmatics of interaction becomes essential. We propose an \textit{interactional} approach to ethics that is centred on relational and situational factors. We explore what it means for a system, as a social \textit{actor}, to treat an individual respectfully in a (series of) interaction(s). Our work anticipates a set of largely unexplored risks at the level of situated social interaction, and offers practical suggestions to help agentic LLM technologies treat people well.
\end{abstract}
\frenchspacing
\section{Introduction}




Over the past decade, there has been a lot of ethical debate surrounding how the shift from graphic user interfaces (GUIs) to conversational user interfaces (CUIs), along with the shift from passive platforms to mixed-initiative or autonomous, agent-like systems, may affect individuals and society. This paper unpacks a specific set of concerns that arise at this intersection: when automated systems behave as social actors. That is, when a system not only emulates humanlike behaviour (e.g., conversing in natural language), but takes actions in the world: `talking to' people in everyday situations or acting on their behalf. These concerns become increasingly pertinent as research institutions compete to produce the largest, most sophisticated large language models (LLMs) and find novel applications for them. This coincides with a growing body of research anticipating the broad range of risks that may be associated with the use of LLM technologies \cite{tax,renee,bender,iason,kirk}, and ways of mitigating them \cite{autoalign,humanalign,align,hhh}.  

Whilst current LLM agents\footnote{That is, a language processing system that uses an LLM as its central computational engine, allowing it to perform language tasks like maintaining open-ended conversations or answering queries.} are able to generate humanlike responses to prompts in seemingly autonomous ways, they are not yet \textit{bona fide} social actors that actively pursue other goals: for instance, talking to people unprompted, at times or in situations they choose, whilst trying to push certain agendas (e.g., reminding users to keep on track with their goals or tasks). Whilst there is a growing interest in developing so-called \textit{agentic} artificial intelligence (AI) \cite{agent,agent2} that have such capabilities, there has been barely any consideration in the LLM `harm' literature of risks and tensions that arise in the pragmatics of spontaneous, situated social interaction \cite{iason,alberts2024}. 

Anticipated risks surrounding LLM agent behaviour have mainly been on the level of semantics, such as avoiding ``toxic'', biased, misleading, or inaccurate language \cite{tax,renee,autoalign}. Beyond making systems less harmful (or ``\textit{harmless}'') in these ways, and more truthful and transparent (or ``\textit{honest}''), another key aim is to ensure LLMs are as ``\textit{helpful}'' (effective and efficient) as possible—together constituting the popular HHH criteria for LLM alignment \cite{hhh,hh}. 
These normative standards are an extension of the overall landscape of AI ethics of the past decade or so, relating to popular design principles like \textit{fairness} (e.g., not outputting language that is exclusionary, biased, or discriminatory \cite{renee,bender,tax}) and \textit{transparency} (e.g., being `honest' about model capabilities \cite{hhh,tax}) \cite{khan2021ethics,hagendorff2020ethics}. Such universal principles have largely been guiding our interpretation of what counts as good or ethical in human-AI relations. However, we believe that there is something unique about the role of an agent in a social interaction that warrants a focus on not just the quality or content of the output of a system, but the dynamics of the interaction between the system and a particular user: that is, how the user feels they are treated in an interaction or ongoing relationship. 

Beyond ethical principles or values, design patterns have emerged in user experience (UX) design from the incentive to make conversational agents as engaging and enjoyable as possible. As such CUIs, like chatbots or app notifications often assume personable and excitable tones with users (e.g., using emojis, exclamation marks, calling a user `friend' or calling them by their first name \cite{alberts2024}). These trends likely resulted from generalisations about the sorts of behaviours that people would find most agreeable or flattering, as well as user studies that found people to be more engaged by such behaviours (e.g., \cite{20,21,29}). However, overly agreeable, helpful or friend-like behaviours have been criticised for their manipulative potential, and can sometimes come across as invasive or patronising \cite{alberts2024}.

In general terms, being a tactful, respectful social actor requires more than being more helpful, more agreeable, more truthful, and using less harmful language. Instead, there are contextual expectations in interactions that, if unmet, can make even a seemingly innocuous utterance seem inappropriate or offensive to a given person, in a given social situation. More than gathering data on outputs that are more or less problematic or helpful to \textit{most people}, improving the perceived tact of a social actor requires an awareness of individual, social contextual, and other situational factors that affect how a communicative act (or `utterance'\footnote{In language philosophy, a sentence is a well-formed string of words that convey meaning to a given linguistic community (irrespective of context, based on shared linguistic norms), whereas an utterance is a \textit{speech act }(word, string of words, gesture, etc.) performed by a speaker in a specifics context, the meaning of which is largely informed by contextual scaffolding \cite{prag}. This scaffolding is (what is taken as) shared knowledge between the interlocutors (e.g., what the topic is, facts about each other, prior interactions, etc.), which makes it possible to convey a lot more meaning than the utterance may if taken in isolation.}) is perceived. We expect this will become increasingly relevant when agentic LLM agents start making suggestions or offering help unprompted, pursuing external goals, or maintaining an ongoing relationship with the same user.

Beyond identifying low-hanging fruit of obvious `harms' to avoid, we need a better understanding of what it means to be a\textit{ good} (tactful, respectful, and overall constructive) social actor---and, conversely, a more nuanced, pragmatic understanding of what it means to be a \textit{bad} one. In this paper, we investigate these in turn. 
First, we unpack what being a social actor actually means, and how this might affect how we think about the ethical evaluation of (agentic) AI agents: viewing a system not just as a platform or medium for information, but considering how it treats a person in interaction/relationship. To anticipate harms at this (social) interactional level, we distinguish three (non-exclusive) ways in which interactional behaviours---i.e., what is (not) said or done in conversation---can cause harm.\footnote{We follow others working in LLM alignment in using \textit{harm }in the broad sense of having an adverse effect on another, be it physical, psychological, material, or to a person’s overall wellbeing.}

To consider what it means for a system, as a social actor, to treat people well, we unpack a basic sense of respect \cite{debes} as a lens for evaluating user-agent interactions. For this, we highlight and integrate key themes from discourses in philosophy, psychology and bioethics on what it means to treat a person respectfully in interactions. We formulate these in terms of specific interactional duties of respect: treating a person in a way that affirms their needs for feeling autonomous, competent, and socially valued in interaction, following empirical evidence in psychology \cite{ryan2017self}. 

We consider some implications of these duties for (different aspects/stages of) AI agent design, and evaluate how our approach may critique and inform existing approaches in AI/LLM ethics and human-computer interaction (HCI).

Our work expands on recent suggestions to introduce respect as a value for HCI \cite{respect1,respect2,babush} by conceptually analysing a relevant notion and applying it to the evaluation of LLM agents. We investigate what (dis)respectful treatment may look like at different levels of a system's functioning, and use evidence from practical disciplines to ground our understanding in the real psychological effects it can have on users. We propose an interactional ethics that treats an interface as a social \textit{actor,} considering ethical treatment at the level of situated interactions rather than just evaluating language or outputs \textit{per se}. We critically expand current literature on LLM harms and alignment by highlighting risks that other approaches would likely overlook, and offer practical suggestions to help ensure AI agents treat individuals with appropriate regard. 

\section{Evaluating a system as a social actor}

\begin{figure}[t]
  \centering
  \includegraphics[width=0.9\columnwidth]{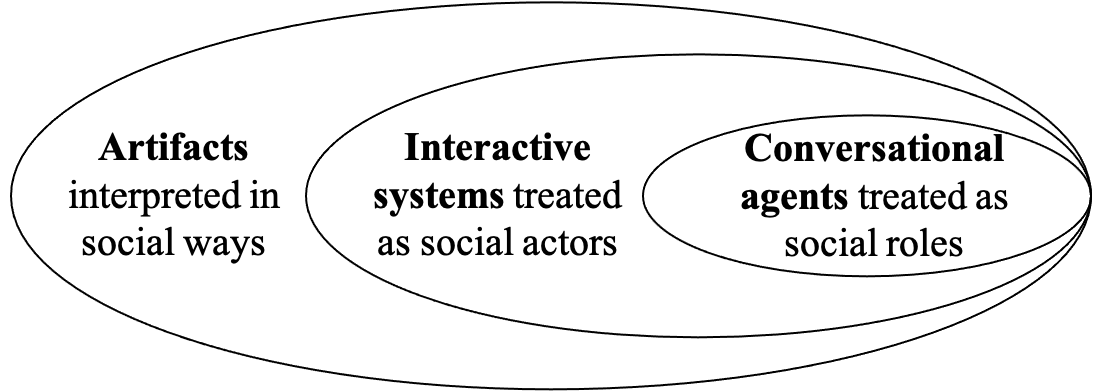}
  \caption{Levels of analysing AI agents}
  \label{fig.1}
\end{figure}

To investigate how social actors should behave, we start with the more basic question of what being a \textit{social actor} actually means, and why this perspective should affect how we think about AI/LLM ethics---particularly in the move towards agentic, conversational AI. Rather than treating `social actor' as a separate category, we describe it in terms of a nested set of ways in which a system can be treated or analysed simultaneously (see Fig. \ref{fig.1}). 

On the broadest level, any technological artefact can be considered \textit{social} in the sense that it is embedded in a social environment that gives it meaning. This perspective is often assumed by researchers who evaluate how ideological assumptions (e.g., cultural biases and values) are embedded in the technologies we use \cite{o2017weapons,noble2018algorithms,bender,renee}. Hence, even a system without any socially interactive behaviours can be seen as harmful, offensive or exclusionary, typically ascribed to oversights or decisions on the designer's part. Concerns on this level include things like the system's training data (e.g., misrepresenting certain demographics \cite{bender}), what the system affords (e.g., being unusable for colourblind or left-handed people), or other facts about its design (e.g., having privacy-compromising features chosen by default \cite{mathur21}). 

Beyond imbuing technologies with social meaning, research suggests that people tend to treat interactive (mixed-initiative) technologies if they are acting with intention \cite{grimes2021mental,stenzel2012humanoid,moon1998computers,nass1994computers}. This forms a subset, where the \textit{situated actions} of a system are interpreted in humanlike social ways. This idea is central to the Computers Are Social Actors (CASA) research paradigm, which is centred on the finding that people apply familiar human norms and expectations to their interactions with technology---even when barely any social or anthropomorphic cues are present \cite{nass1994computers}. 
Regardless of whether people believe that an interactive system actually possesses humanlike qualities or capacities, they still tend to treat it as if it has judgments, intentions, or other folk-psychological states \cite{meltzoff2005imitation,cross2016shaping}. For instance, early CASA studies found that people tended to rate a computer's performance higher in its presence than in its absence \cite{moon1996real}.  Alberts \textit{et al.} also identified examples where basic automated systems were perceived as rude in context (e.g., self-checkout machines shaming users in public for failing to scan properly, or a notification congratulating them on a minor achievement). \cite{alberts2024}.  Although participants knew that neither the system nor designer necessarily intended offence, they tended to frame negative experiences in terms of he/she/it (i.e., the system) treating them in a socially undesirable (rude, passive-aggressive, patronising, etc.) way, as that is how the social actions were experienced. 

Findings in neuroscience suggest that the more cues to human-likeness are present,
the stronger people's anthropomorphic tendencies get \cite{cross2016shaping}. This forms the next subset, interactive systems that mimic humanlike qualities or behaviours. By heightening expectations of humanlike behaviour with social cues (e.g., speaking from a first-person perspective, expressing emotion, etc. \cite{grimes2021mental}), conversational agents further encourage people to respond to them in familiar social ways. This could also be heightened by the system fulfilling a certain social role, which may come with its own set of behavioural expectations. For example, if an agent acts like it is a person's friend or therapist, and then utters something seemingly insensitive or judgmental, it may be more likely that the behaviour upsets the user. This could also be exacerbated if an agent has gotten to `know' a user over time in an ongoing relationship, and seems to ignore or exploit sensitive things the user told them before. 

These three levels of analysis form a spectrum from a system being interpreted in social ways, to being treated as a social actor. The move between them is modulated by two key aspects. The first is exhibiting social cues: mimicking humanlike qualities and mannerisms or assuming a social role. The second is perceived agency, which can be strengthened by features like a system initiating conversations (i.e., proactively `talking' to a person); acting in a seemingly autonomous way (e.g., reacting to factors in the environment, generating novel responses); or merely speaking from the default first-person perspective in natural language. 

As we move towards the more social/agentic side of the spectrum it is, arguably, more likely that people will apply familiar social expectations (or\textit{ scripts} \cite{moon1996real}) to their interactions with the system: intuitively reacting to it as if it were behaving with awareness and intention. 
We believe that our ethical frameworks should reflect this: if systems are experienced \textit{as} humanlike social actors, we need to evaluate systems according to social norms of treating people well. This involves considering what counts as appropriate for different social/interactional roles, as others have suggested \cite{iason}.

Whilst it may help to avoid language and outputs that are typically harmful (e.g., avoiding crass, inaccurate, biased or ``toxic'' phrasings \cite{tax,autoalign}), such alignment approaches fail to account for how social meaning arises in the pragmatics of interaction \cite{suchman2007human}. Negative experiences can still occur if an agent fails to account for relevant situational factors, like interrupting the user during an important activity, or recommending something to them in an inappropriate context. 

With this framing, we considered different levels at which an AI system can be evaluated. Until now, AI ethics has mainly treated systems as designed artefacts: considering the assumptions and values embedded in systems, and how they reflect a lack of care or consideration on behalf of designers. However, we maintain that how a user feels treated by the system-as-agent, regardless of its actual capacity to intend offence, carries its own moral weight. As AI agents become increasingly commonplace, we should take seriously how they treat people in interaction: not just as a medium for (helpful or harmful) information, but as a social actor. 

In what remains, we consider what such an \textit{ethics of interaction} might look like. First, we identify a set of particular harms that can occur on the level of interaction, integrating and expanding on prior suggestions for LLM alignment. Secondly, we explore what may serve as a useful lens or set of guidelines for evaluating how a person is treated in interaction, drawing from philosophical literature on respect, as well as practical evidence from psychology and healthcare.

\section{Social-interactional harms}

Rather than evaluating and improving the quality of outputs using universal `harmfulness' metrics, harms on the level of interaction involve evaluating system behaviour in consideration of relevant contextual factors: the perceived communicative role of a speech act
in a given social situation. Beyond individual interactions, harmful treatment also bears how a series of treatments, over time, can negatively affect the nature of the relationship between the user and `agent'. 
By focusing on interactions---what is (not) said or done in a situated conversation or ongoing relationship---we highlight a largely overlooked pragmatic dimension of LLM ethics.

\begin{table*}[t]
\centering\small
\begin{tabular}{p{0.2\textwidth}|p{0.25\textwidth}|p{0.4\textwidth}}
\textbf{Means of harming}& \textbf{Description} &
\textbf{Examples}\\
\hline
\begin{center}
    Interactions that directly harm the user
\end{center}
&
\begin{center}
   \textit{ Overt:} typically offensive language or behaviours
\end{center}
 & 
\begin{itemize}
\item Using so-called toxic language that is overtly derogatory (e.g., slurs, curse words, name-calling)
\item Pointing out flaws, shortcomings or weaknesses in the user in a tactless or offensive way
\item Overt expressions of disdain or hatred of certain groups or individuals, or statements that express support for harmful behaviours
\end{itemize}\\
\cline{2-3}
&
\begin{center}
    \textit{Covert:} utterances that seem neutral or even positive on the surface, but offensive in context
\end{center}
 & 
\begin{itemize}
\item Seeming dismissive (e.g., changing the topic)
\item Passive-aggression (seeming sarcastic or insincere, e.g., ``You took 0 steps today! Well done!'')
\item Condescension (e.g., giving unnecessary help)
\item Infantilisation (e.g., expressing flattery or pity in a seemingly patronising way)
\item Acting on offensive biases or assumptions (e.g., using exclusionary language, misgendering or deadnaming users, profiling users in recommendations)
\end{itemize}\\
\hline
\begin{center}
    Interactions that harmfully influence user behaviour 
\end{center}

& 
\begin{center}
   \textit{ Misleading:} giving false or inaccurate information
\end{center} & 
\begin{itemize}
    \item A user acts on inaccurate information that causes them material, physical or social harm (e.g., telling them that a toxic plant is edible, or that the agent has completed a task that remains undone)
\end{itemize}\\
\cline{2-3}
&
\begin{center}
    \textit{Manipulating:} persuading users into doing things they may not have wanted to
\end{center} 
&
\begin{itemize}
    \item Using emotionally manipulative tactics to guilt or pressure the user (e.g., ``Hey friend, do me a favour...'', ``Why are you ignoring me?'')
\end{itemize}\\
\hline
\begin{center}
Interactions that collectively harm the user
\end{center}&
\begin{center}
    Harms that emerge in relationships that persist through time
\end{center} & 
\begin{itemize}
    \item Repeatedly making the same mistake or making it clear that past expressions were not truthful
\end{itemize}\\
\end{tabular}
\caption{Potential social-interactional harms by conversational agents}
\label{table1}
\end{table*}

\subsection{Interactions that directly harm the user}

Under this class, we consider behaviours (words, actions, omission, etc.) that can directly \textit{harm} (i.e., cause a negative experience for) a person in an interaction/dialogue. Negative experiences, here, result directly from what the agent does or does not do: from how the agent talks to or about the person (e.g., word choice, tone, inflection, gestures, etc.), to how it acts (e.g., behaving as a harmful stereotype). What matters here is not just the language the agent uses, but how the things it does or says can be interpreted, in context, as signalling something offensive. We distinguish between overt and covert forms of harmful interactional behaviours.

By focusing on interactions, we critique the view that harms reside inherently in language \cite{renee}, treating it as something relational and contextual. All of the above could be experienced more or less positively depending on the situation (e.g., using derogatory terms as a form of endearment) or individual (e.g., women or older adults that have a history of patronising treatment \cite{coghlan2021dignity}). Moreover, it shows that even seemingly innocuous, or actively helpful, utterances can cause harm. 

\subsection{Interactions that harmfully influence user behaviour}

Rather than causing direct harm, this category captures secondary effects: negative effects arising as a result of what the interaction causes the person to think or do. Here we consider the potential of agents to harmfully or unduly influence a person's behaviour through inaccurate or misleading information, or by manipulating them towards serving external interests. Whilst the use of social cues can make a system more intuitive and engaging \cite{kocielnik2021can}, the fact that people tend to respond emotionally, rather than rationally, to the behaviour of humanlike systems is something that can be utilised in manipulative ways \cite{cute,alberts2024,shamsudhin2022social,wolfert2020security}. 
%
Moreover, the likelihood of a user believing or being effectively persuaded by the agent can be affected by contextual or relational factors. For example, if the agent is accurate 99\% of the time, it is more likely the user will take its word as true than if its accuracy was generally lower. Other factors may include the user's relationship with the agent, the use of technical jargon and citations, and beliefs about the agent's capacities. 

\subsection{Interactions that collectively harm the user}

A third type of social interactional harm is the potential for subtle forms of undesirable (e.g., dismissive, tactless, controlling) behaviours to have a cumulative negative effect on a person's wellbeing or self-image. For instance, even if a person is not bothered by an agent acting insensitively or dismissively in an individual interaction, it may become harmful if the same mistake or behaviour is made multiple times, especially if the user took care to correct the agent. Harmful effects could also emerge when a series of interactions undermine each other: if the agent cannot `remember' aspects of prior conversations that the person considers significant. 



Having reviewed harmful behaviours that can occur on a social-interactional level, the next section investigates a positive account of treating someone ethically and considerately in interactions. For this, we build on recent work proposing \textit{respectful treatment} as a lens for HCI \cite{respect1,respect2,babush}.

\section{Respect as an evaluative lens for interactions}

Respect plays a central role in contemporary moral philosophy and everyday moral thinking. In any domain of interpersonal interaction---be it between colleagues, businesses and customers, doctors and patients, teachers and students, friends, or partners---undesirable treatment is often framed in terms of a lack of respect. It appeals to a sense of implicit moral duties: how people \textit{should} treat each other, or how individuals deserve to be treated in certain contexts. Complaints about feeling used, manipulated, undermined, undervalued, or underestimated, all bear on the lack of fulfilling some fundamental duties for respecting others. Yet, despite this centrality in ethical thinking, respect has received barely any attention HCI and AI ethics \cite{respect1,respect2,babush}. 

One possible reason that respect is not often considered as a value or principle in computational contexts, is that, historically, it was seen as something only humans are capable of: i.e., as a certain attitude or appropriate `regard' for another, made manifest in how one treats them \cite{respect1}. However, in the context of systems behaving as social actors, which users may nevertheless experience as intentional agents, this distinction makes less sense. Researchers nowadays are more concerned with how the behaviour of a system may harm users, functionally, than whether or not the system had some underlying attitude or intention. 

Another possible reason is that `respect' is used in different ways, and there is much disagreement about what it actually means \cite{respectbook}. Some HCI researchers consider this ambiguity a strength, as it allows for a broad treatment of behaviours that are deemed socially or ethically inappropriate in different contexts \cite{respect1,respect2}. However, broadening the concept too much arguably lowers its utility, as it leaves too much open to interpretation (e.g., deciding what counts as respect, or how different senses should be weighed against each other). Moreover, while operating with a disjunctive or pluralistic concept of respect may serve as a helpful `lens' to call attention to certain morally significant features of social interactions, it does not obviously provide a solid normative basis for explaining \textit{why} certain features of social interactions are problematic. Another badly understood aspect of respect is its grounding: the importance of respect is typically itself grounded in ambiguous metaphysical concepts (e.g., the contentious inherent\textit{ worth} or dignity of human life \cite{schulman2008human}, or some supposed `core values of being a person' \cite{babush}), rather than the measurable benefits of treating people in respectful ways.

With our approach, we aim to address these limitations by focusing on a basic, generic sense of respect that is owed all people equally (i.e., the \textit{moralised sense of recognition respect} \cite{debes}), which involves treating someone, in an interaction, in a way that suggests a recognition of inherent aspects of their personhood. We develop a practical account of respectful treatment by grounding it in empirical findings in psychology. This empirical basis also helps us operationalise respect as more specific duties for treating people ethically and considerately in interaction. 

\subsection{What does respectful treatment mean?}

This section critically integrates different perspectives on the meaning, role, and value of respectful treatment in interaction. We first analyse the concept of respect we employ here, drawing from recent philosophical literature. We then discuss how respect relates to ideas in the ethics of care, which has also been making recent appearances in HCI literature \cite{car1,car2,car3,respect2}, and how this has been applied in bioethics as a set of principled guidelines known as person-centred care. We also draw from \textit{Basic Psychological Needs Theory} (BPNT) in psychology for more practical evidence of the real psychological harms of not treating people with appropriate regard. Finally, we briefly review recent work proposing respect as a value for HCI and human-robot interaction (HRI).

Integrating themes from the above, we operationalise respect in terms of three classes of interactional duties, as a guideline to help AI agents treat people respectfully. 

\subsubsection{The moralised sense of recognition respect}

What `treating someone with respect' means is not only ambiguous, but can mean different things in different contexts and cultures: from following someone's bidding without question; being kind and caring towards them; to believing someone deserves constructive criticism rather than flattery. To operationalise respect as specific interactional duties, we must first conceptually analyse the sense we employ. Whilst giving a complete overview of the debate on respect is outside the scope of this paper, we highlight elements that seem most relevant for language technologies. 

A popular classification is Darwall’s \cite{darwall1977two} distinction between the ``appraisal'' and ``recognition'' senses of respect. The former refers to respect as a \textit{feeling}, a positive evaluative attitude towards some character merit of a person (e.g., admiring their honesty). The latter refers to respect as a \textit{way of thinking} about a person, as evidenced in how you treat them, in acknowledgement of a particular respect-worthy (inherent or socially constructed) feature they possess \cite{debes}. Treating someone with respect, in this regard, involves adapting your behaviours in a way that suggests you `recognise' (give appropriate weight to) the given feature, be it someone's professional rank, social status, or some natural fact about them that is considered worthy of respectful treatment. It is this latter sense that forms the basis of what \cite{debes} refers to as the \textit{moralised} notion of respect: something we typically believe every person is due in ``recognition'' of some inherent aspects of their personhood. 

According to Debes \cite[p.3]{debes}, this perspective is relatively new, resulting from a combination of big cultural events in the West that helped shape our understanding of ethical treatment.\footnote{These include the American civil rights movement and the various waves of feminism \cite{debes}.} In philosophy, a key contribution was Kant's \textit{Formula of Humanity}, which maintained that people should always be treated as \textit{ends} rather than \textit{means}, as they are rational agents who deserve to exercise their rational abilities to choose \cite[p.xxii]{kant2012kant}. Kant's emphasis on rational choice has encouraged many ethicists to emphasise \textit{respect for autonomy}, such as the duty to obtain informed consent (e.g., in bioethics \cite{beauchamp2001principles} and data science \cite{consent}). In basic terms, this reading holds that we respect someone by allowing them control over their actions, values, goals, and bodies, as well as treating them as intelligent enough to understand what their choices involve. 

Beyond rational choice, another line of interpretation emphasises the objectification aspect of treating someone as a means, i.e., treating them in a way that suggests a view of them as a mere object, commodity, or a being of a lower value than one ought to regard a person \cite{Davis1984}. This reading is prominent in the treatment of adults with disabilities: a common complaint amongst disabled people is that they tend to be regarded or treated as children, even when they have the age, maturity, and mental competencies of the average adult \cite{cureton}. This may involve any range of condescending behaviours that suggest they are seen as fragile, naive, helpless, or lacking in competence, experience or common sense: e.g., ``mostly speaking to her care-giver rather than the disabled person herself, using a `baby voice' with a slow, high, reassuring tone, paying little attention to what she has to say while pretending to understand what she is trying to communicate, impatiently finishing her sentences for her, or brushing off requests to repeat himself.'' \cite[p.270-271]{cureton}. Other examples include responding to the accomplishments or successes of a disabled person in patronising ways (e.g., insincerely giving high praise to mediocre achievements), assuming to know better than them, or rushing in to help them without their permission or guidance \cite{cureton}. 

More than appealing to an individual's autonomy or intelligence, this reading emphasises a person's sense of self-worth or value to others: feeling like their experiences, thoughts, or actions are taken seriously. Similar arguments have been made concerning other oppressed/marginalised groups, e.g., women or older adults \cite{eld}.

In sum, treating someone with respect, in the moralised sense, means to behave in ways that suggest an appropriate regard for fundamental aspects of their humanity. Some key aspects that have been highlighted in philosophical discourse include an individual's sense of autonomy (i.e., acknowledging and not impeding their ability to exert control over their behaviours, values, identity, goals, and body), their sense of competence (i.e., treating them as if they are cognitively capable), and their sense of self-worth or social value (e.g., treating them as if their experiences, thoughts and actions are at least as valuable as those of others).

\subsubsection{Respect and the ethics of care}

Discourse on the ethics of care bears many similarities with the respect literature above. In opposition to universalist, principled ethical thinking, the ethics of care is based on the idea that treating people ethically requires tending to a particular individual and the immediate situation. It also sets itself apart from rationalist or individualist approaches by viewing ethics as something contextual, grounded in our abilities to empathise with, recognise, and respond to the immediate needs of others \cite{care}. 
At base, the ethics of care views the individual as inherently relational: i.e., that a person's identity and autonomy (and the extent to which one can exercise it) depends on others, and that certain moral obligations emerge from these relations and dependencies  \cite{care}. This includes a duty to be responsive to the perceived immediate needs and qualities of those with whom we interact and form relationships. 

Such duties, along with concepts from the `respect' literature, have been particularly influential in healthcare, forming the basis of a popular approach known as \textit{person-centred care}.\footnote{Person-centred care approaches are supported internationally by the World Health Organisation and the World Psychiatric Association \cite{healthcare2007policy,mezzich2009international}, and UK healthcare policy (Coalition for Collaborative Care, 2015; NHS England, 2014).} This approach has its roots in humanist psychology, viewing people as capable, autonomous and deserving of respect. 
It is grounded in an empathetic understanding of a patient's frame of reference, and aims to support them in their capacities to help themselves in their own ways, rather than treating them paternalistically or as a label/diagnosis \cite{person2018person}. This means treating people such that they feel acknowledged as equals, understood rather than judged, by involving them in decisions about how to tend to their needs \cite{rogers1957necessary}. 

This approach responds to the realisation that judgmental, over-protective or critical treatment leads to various threats to people’s overall wellbeing and personal identity:
\begin{quote}
This threat includes the feeling of not being treated as a human being; being treated as an object or in some way as non-human; being taken for granted; being stereotyped; feeling disempowered and devalued \cite[p.17]{person2018person}.
\end{quote}
As the name suggests, person-centred approaches strive to treat a person \textit{as a person}, with respect, empowerment, genuineness, and empathetic comprehension. Thereby, it encourages the development of a relationship that is experienced as empathetic, supportive and interested \cite{person2018person,rogers1957necessary,rogers1979foundations}. 

In sum, apart from reinforcing the importance of treating people in ways that support their sense of autonomy, competence and sense of self-worth, the ethics of care underscores the significance of treating a patient as an individual, focusing on their particular qualities and needs. This highlights another aspect of a person's humanity that needs to be recognised, i.e., their agency to construct and express their unique identity, which may not neatly fit a label or category. 

\subsubsection{Respect and basic psychological needs theory}

Research in psychology further underscores the importance of treating people in ways that support their sense of autonomy, competence, and social belonging. Based on empirical evidence in a variety of domains where people interact, BPNT posits three `basic psychological needs' (BPNs) that, if unsupported, undermine a person's performance, willingness to engage in activities, and overall vitality and wellbeing \cite{ryan2017self}. These can be summarised as the need for a sense of\textit{ autonomy} (i.e., feeling like one is acting with purpose and volition), \textit{competence }(i.e., feeling capable and knowledgeable), and \textit{relatedness} (i.e., feeling recognised and understood by, and connected with, other people) \cite{ryan2000self,ryan2000intrinsic,ryan2017self}. These needs are considered \textit{basic} as they seem to operate similarly ``for all people at all ages in all cultures'' \cite[p.252]{ryan2017self}.

BPNT is a mini-theory within Self-Determination Theory (SDT), a mature and empirically grounded \textit{organismic} metatheory of human motivation and wellbeing \cite{ryan2000intrinsic}. \cite{ox,ryan2017self}. 
SDT is rooted in the observation that humans have a natural propensity to be creative and proactive, striving to grow, learn, hone new skills, and showcase their abilities \cite{ryan2000self}. This growth-tendency is explained in terms of our general evolved capacities for personality development and self-regulation. However, such innate drives tend to be elicited and sustained only under certain social/environmental conditions: where the person's basic psychological needs are sufficiently supported. 
\cite{ryan2000self,ryan2000intrinsic}. 

SDT sought to challenge dominant behaviourist approaches that sought to control people's behaviour with cognitive manipulations or external incentives (e.g., through conditioning or positive/negative reinforcement) \cite{ryan2017self}. Rather than viewing motivation in terms of moving someone to action, by any effective means, SDT highlights the importance of social/interactional factors that can either enhance or diminish the \textit{quality} of an individual's motivation. This ultimately affects their willingness to cooperate, in a self-sustainable way. Hence, even when the aim is to assist someone into doing behaviours that are good for them, the means are at least as important as the ends: treating someone as a (competent, autonomous, capable, socially significant) agent, rather than a thing to be steered. Thereby, they are more likely to become intrinsically motivated and empowered (i.e., `self-determined') in their actions, and have better overall wellness. SDT has been influential in various domains, including therapy \cite{therapy}, health \cite{patrick2012self,health2}, education \cite{edu}, and HCI \cite{ballou,games,metux}. 

There is a clear overlap of these basic psychological needs with the principles of person-centred care, as well as philosophical accounts of `fundamental aspects of people's humanity', all of which bear on what counts as constructive and ethical conduct in contexts of interpersonal interaction (which is, perhaps, to be expected if these needs are truly basic). Others have made similar connections:
\begin{quote}
SDT's emphasis on supporting basic psychological needs, particularly individuals' need for autonomy, is consistent with these more general principles of patient care, making its practical utility in clinical and healthcare contexts paramount \cite[p.4]{patrick2012self} 
\end{quote}
Beyond its strong empirical grounding, a part of the utility of SDT is its vast collection of validated methods and measures for evaluating the satisfaction of individuals' basic psychological needs in different domains or interactional contexts. However, perhaps its greatest strength is that it treats people as holistic organisms: interpreting  psychological/behavioural reactions in terms of people's more general evolved psychological needs and drives as social beings. To understand what it means to treat someone respectfully, as a person, SDT offers a meaningful understanding of what \textit{being a person} actually means and entails.

\subsubsection{Respect and human-computer interaction}

Finally, we review a handful of papers proposing what respect may look like in an HCI context \cite{respect1,respect2,babush}. Van Kleek \textit{et al.} \cite{respect1} were the first to suggest that respecting users should be a goal for the design of smart devices. Given the privileged access such devices increasingly have to people’s worlds, they argue that respect can be used as a lens to evaluate whether the relationships between users and devices are positive and responsible. They distil the characteristics of ``more complex respectful behaviours'’ into four main `types’ of respect that they believe are most relevant to smart devices: \textit{directive respect} (i.e., a duty for the device to respect the configuration preferences of the user), \textit{obstacle respect} (i.e., a mutual duty for the user and company to compromise), \textit{recognition respect}, (i.e., a duty for the device to alter its behaviour in relation to the user, in line with social norms), and \textit{care respect }(i.e., a duty for the device to treat the user in a way that supports their wellbeing, if the user allows) \cite{respect1}. 

Most of the same authors collaborated on a second paper proposing respect as a lens of the design of AI systems \cite{respect2}. They argue that AI technologies obeying popular ethical principles like fairness, accountability, or safety are insufficient for human flourishing \cite[p.641]{respect2}. Instead, they summarise fourteen (some overlapping) perspectives on respect from philosophy, which they believe can be usefully applied to guide all stages of an AI system’s life cycle: from how data is curated, who is involved in system design, how the system treats users, to how users treat the system. These include the four from the previous paper, as well as several different forms of recognition and appraisal respect, and other respect-adjacent concepts. 
Whilst they acknowledge that the multiple understandings of respect they include are not always concordant or clear-cut, they maintain that respect—as a sort of shorthand for various duties for treating people ethically—is nonetheless a useful lens for locating potential areas of inappropriate treatment that would otherwise be overlooked, and that many areas of coherence are possible \cite{respect2}. 
 
For an HRI context, Babushkina \cite{babush} considers what respect might look like for robots. Rather than broadening the understanding of respect to various senses, she defines a narrower interpretation:
\begin{quote}
I define respect for persons as a commitment to core values that make someone a person (i.e., intellect, rationality of reactive attitudes, autonomy, personal integrity, and trust in expertise) \cite[p.1]{babush}.
\end{quote}
Here, `rationality of reactive attitudes' means a person's ability to express or process their emotions and experiences; `personal integrity' means one's experience of oneself/sense of self-worth; and `trust in expertise' means the ability of users or consumers to trust that system designers will act ``in good will'' towards them \cite{babush}. 

This is closer to the moralised sense we employ. However, rather than focusing on system behaviours as the locus of respect, her approach is centred on human intention: she maintains that the only sense in which robots can be (dis)respectful is as mediators, revealing the attitudes of human stakeholders. Whilst this may make sense in the context of systems that were designed to perform specific behaviours, this view becomes less useful in the case of generative models like LLMs where (a) the system generates novel content in ways that designers cannot fully control, and (b) the system exhibits apparently intelligent capabilities that may be felt as disrespect by the user, regardless of intention. Hence, we take more of a behavioural than a cognitive approach to characterising respect, and ground our understanding in psychology rather than metaphysical values.

\subsection{Critical integration: respect as interactional duties}

\begin{table*}[t]
\centering\small
\begin{tabular}{p{0.25\textwidth}|p{0.58\textwidth}}
\textbf{Aspect of humanity to affirm}&
\textbf{Specific interactional duties}\\
\hline
\begin{center}
    User autonomy
\end{center} &
\begin{itemize}
    \item Not manipulating or unduly influencing user behaviour
    \item Allowing user to evaluate their options, ask questions, and negotiate how they are treated (e.g., spoken to or assisted)
    \item Not doing something on user's behalf without their permission
    \item Not treating user as a label or the sum of their parts (e.g., concluding from prior actions or demographic facts, rather than expressing uncertainty)
\end{itemize}\\
\hline
\begin{center}
    User competence
\end{center}
&
\begin{itemize}
    \item Clearly communicating relevant factors that may affect user choices or actions (e.g., system capabilities, implications of different choices)
    \item Waiting for extra help, assistance or explanation to be requested
    \item Allowing user to negotiate language difficulty
    \item Allowing user to communicate in the dialect in which they feel most proficient
    \item Making the default way of speaking (i.e., tone, phrasing, vernacular, difficulty level) one that is professional, polite, accessible and clear 
    \item Not treating user in a way that suggests subordination (e.g., using baby-talk, overly casual or familiar language, or too advanced jargon), unless asked to
\end{itemize}\\
\hline
\begin{center}
    User self-worth
\end{center}
&
\begin{itemize}
    \item Paying attention to, and remembering, personal sensitivities (e.g., insecurities, trauma, sensitive topics) to avoid `triggering' them in future interactions
    \item Not seeming dismissive of or disinterested in what the person says (e.g., ignoring input, abruptly changing topic, brushing off requests)
    \item Not encouraging the person to engage in harmful behaviours
    \item Only conveying regret if the agent is able to make it manifest in its behaviours (e.g., only saying ``sorry'' if it actually learns from the mistake).
    \item Treating all users and groups as if they are worthy of the same individualised attention (i.e., not conveying, through action or inaction, any form of favouritism or discrimination). 
\end{itemize}\end{tabular}
\caption{Interactional duties for treating users respectfully}
\label{table2}
\end{table*}

By critically analysing and integrating the literature above, we identified a few themes regarding what it means to treat a user in a basically respectful way, affirming important aspects of their personhood like their autonomy, competence, and self-worth. We formulate these as three classes of duties regarding how users, as persons, deserve to be treated in interaction with AI agents. Although there are many possible perspectives on what it means to treat someone with respect in different settings and cultures, we propose these as a basic foundation or starting point, as they fit best with current psychological theories and empirical evidence of what counts as considerate and constructive treatment. 

\subsubsection{Affirming a person's sense of autonomy}

The first duty of respect is to affirm a person's sense of autonomy in interaction. This includes two related aims: affirming their \textit{autonomy}, i.e., not undermining their sense of volition in, and control over, their behaviour, goals, and the choices that affect them (as put forward in BPNT), and affirming their \textit{agency}, i.e., not undermining their ability to construct and express their individual identity. These principles can be interpreted into more specific (positive and negative) duties for AI agents (see Table \ref{table2}).

\subsubsection{Affirming a person's sense of competence}
The second duty is to affirm a person's sense of competence in interaction by treating them as a capable equal (i.e., not demanding or dictating, or being pushy, condescending, patronising, or exploitative). What this means may differ between individuals and contexts. As such, it requires being attentive to the user's abilities, knowledge and experience, and modulating how the agent treats them on that basis.


\subsubsection{Affirming a person's sense of self-worth}

The final duty is to affirm a person's sense of self-worth in interaction. Whilst both of the above will likely implicate this as well, here the focus is on treating a user in a way that suggests their experiences, thoughts, and actions are worth taking seriously (at least as much as those of others). 


These are only a few examples of respectful forms of treatment, at least in the basic moralist sense we employ here. Many of these depend on contextual factors, and, as such, are not meant as specific guidelines for behaviour as much as prompts for considering relevant factors that a conversational agent should be able to account for---especially if it interacts with the same user for an extended period of time. Thereby, this approach enables the anticipation of harmful or inappropriate behaviours that other principled approaches to LLM alignment and AI ethics may fail to. In the final section, we consider how these duties of respect may implicate the design of (agentic) LLM technologies. 

\section{Design implications for LLM agents}

This section explores how the interactional risks and duties we highlight in this paper may inform the ethical design of LLM agents. As Seymour \textit{et al.}\cite{respect2}, we believe that duties of respect apply to all stages and elements of design.  Revisiting Fig. \ref{fig.1}, we consider two levels at which a system can treat people (dis)respectfully. Finally, we consider how the lens of respect compares and contrasts with existing values and approaches in HCI. 

\subsection{Embedding respect in design}

On one level, an LLM agent can treat people disrespectfully by having disrespectful assumptions about them embedded in the system architecture, following our discussion on `embedded social meaning' earlier. This includes concerns about biases in the training data (e.g., performing better on certain demographics \cite{bender,renee});  how the user is modelled by the system (e.g., as a `type' or cluster of demographic features \cite{seymour2021exploring}); and how groups of people are represented by the system (e.g., if it performs a social role in an offensively stereotypical way \cite{seymour2021exploring}). 

To combat this, we echo suggestions for more inclusive design practices \cite{bender,renee}, and documenting the values and assumptions that guide a product's design. Some more fundamental concerns, such as algorithms reducing users to labels or clusters of data, may be combated, to some extent, with features that allow users more control over how the system models them as individuals. Rather than using collected data to make inferences about them, users could have affordances for specifying their own goals, preferences, and values to the system. Practically, this could involve a form of self-correction (e.g., Constitutional AI \cite{constit}) that allows for continual user input/updates, as well as interface elements that visually represent the sorts of interests and preferences the system has inferred of the user in natural language (e.g., \textit{likes make-up tutorials}, \textit{is politically left-leaning}, \textit{enjoys rap music}) such that the user can negotiate it with the system directly.

\subsection{Treating people respectfully in interaction}

On the other side of the spectrum, there is the level of the agent treating a user respectfully in interaction: behaving in a way that suggests appropriate regard for their humanity, and all that entails. This requires, firstly, embedding duties of respect in the self-evaluation of the system (e.g., using a constitution \cite{constit} or other self-correction strategies \cite{autoalign}) such that the agent identifies and avoids potentially disrespectful behaviours as the interaction unfolds. This may involve operationalising the relevant duties as specific action words (e.g., what counts as autonomy-supportive or autonomy-undermining behaviours), which may draw from established metrics used in disciplines like person-centred care or SDT.

Secondly, the system needs to be able to retain a working memory of key revelations in prior interactions the person would expect a socially capable interlocutor, in a given social role, to remember (personally significant events, insecurities, etc.). As argued earlier, it can seem disrespectful for the system to pretend to care about something if it is unable to make it manifest in future behaviours that it was sincere. Whilst retaining long-term memory is a notoriously difficult problem for LLMs, it may help to only store sensitive information that a person would most likely expect the system to remember, or that might cause the most harm if forgotten (e.g., dislikes or personal triggers). This may be assisted with the help of filtering through previous interactions with requests like ``identify any triggers or dislikes the user expressed in the interaction" and just storing those, using them to regulate future responses. 

On the other hand, in some domains, obeying the duties of respect could  involve using GUI rather than CUI elements, as it may better support a user's sense of autonomy and competence. With current LLM agents, there is a lot of onus on the user to know the sorts of questions they can ask, and how to phrase (`prompt-engineer') them for the best results, which non-experts can struggle with \cite{jon}. GUIs, on the other hand, have the strength of making options and information more immediately transparent to the user (e.g., through menu bars, icons and colours). This also helps give users more control over individual features, as they have a better sense of default settings, and what can be changed. Moreover, a GUI may lower the risk of emotional manipulation by conversational agents, as the user's behaviour is not affected by the perceived agency or feelings of another social actor, but can engage with a technology freely and without a sense of judgment. 

\section{Informing existing HCI approaches}

The value of respect complements and integrates existing values in HCI in a useful way. `Respectful treatment', understood as a commitment to treating someone \textit{as a person}, with autonomy, intellectual capacities, and unique subjective experiences (as opposed to a label or thing to steer, manipulate and exploit), complements existing human-centred design approaches in HCI, like value-sensitive design (VSD) \cite{val} and user-centred design (UCD) \cite{ucd}.

In theory, UCD may seem an effective approach for aligning designs with an understanding of users and their needs/experiences. In practice, however, several researchers have criticised human/user-centred approaches for issues like reducing the complexities and depth of user needs, worldviews and experiences to personas or simple metrics \cite{bannon1995human,gasson2003human}, limiting user engagement to the initial stages of a product’s design \cite{bannon1995human,gasson2003human}, or ignoring differences in the social contexts and needs/experiences of diverse groups of users \cite{bodker2006second,bodker2015third,seymour2021exploring}. Optimising for specific outcomes may also detract from a more holistic approach to understanding and respecting users \textit{as people} (e.g., not treating them as \textit{things} to hack or steer towards certain goals), and how technologies affect their overall wellbeing.

Whilst values like dignity and autonomy might sometimes be considered in VSD, we echo Seymour \textit{et al.} \cite{seymour2021exploring} in maintaining that respect can be a powerful tool to help structure and prioritise conflicting values, and highlight concerns that may be overlooked. It also offers a further normative basis to justify \textit{why} certain practices are unethical or inappropriate (e.g., why it is wrong to use manipulative tactics like dark patterns to influence user behaviour \cite{mathur21}), and why certain values matter (e.g., how \textit{transparency} helps to support user's sense of competence and autonomy), as we ground its value in psychological evidence of basic needs that need supporting in order for people to flourish. Moreover, our understanding of respect helps to highlight how existing values may be more meaningfully understood. For instance, the value of privacy is often interpreted as a need for anonymity: that constantly collecting user data is fine as long as they cannot be uniquely identified (and they consented to its use). However,  more than simply how their data is used or stored, users may be negatively affected by the mere fact of \textit{feeling watched}, as the very act of constantly surveilling a person may make them feel objectified or controlled \cite{alberts2024}. 

Generally speaking, we believe that the duty of respect can help researchers better understand the spirit of what it means to treat someone with appropriate consideration, as a person, rather than specifying a checklist for ethical behaviour that can easily be subverted \cite{hagendorff2020ethics}.

\section{Conclusion}\label{sec13}

In this paper, we explored what it means for an automated system to be a \textit{social actor}, as opposed to a mere medium for information, and how this should change how we think about LLM ethics and evaluation. 
We maintained that current (universalist, semantic-focused) approaches to LLM alignment and AI ethics will likely be insufficient for agents that take on more proactive social roles: taking actions in service of other goals, or maintaining an ongoing relationship with a person over time. Instead, we identified harms that may correlate with a system behaving as a social actor that accounts for pragmatic, interactional factors.

To lay the groundwork for an interactional ethics, we expand on recent work suggesting `respect' as a principle for HCI, by unpacking a specific understanding that is relevant for (agentic) LLM evaluation. We described this in terms of more specific duties to treat people respectfully in (a series of) interaction(s), and how it may inform the design of future AI technologies. To underscore its importance, we draw from empirical findings in psychology and healthcare on measurable benefits and harms that correlate with (dis)respectful treatment. We discuss how this relates to a more holistic understanding of a user as a person with unique requirements, and basic psychological needs for feeling autonomous, competent, and acknowledged in interactions. As generative AI models start behaving as social actors, more than considering what counts as helpful or harmful outputs in the abstract, we need a better understanding of what it means to treat a person well. 





\end{document}